\documentclass[10pt,twocolumn,letterpaper]{article}

\usepackage[]{wacv} 

\usepackage{graphicx}
\usepackage{amsmath}
\usepackage{amssymb}

\usepackage{caption}

\usepackage[numbers,sort&compress]{natbib}

\usepackage{booktabs}
\usepackage{pgfplots}
\usepgfplotslibrary{fillbetween}
\pgfplotsset{compat=1.18}
\usepgfplotslibrary{groupplots}
\usepackage{array}

\definecolor{cElect}{RGB}{31,119,180}
\definecolor{cConsi}{RGB}{214,39,40}
\definecolor{cOnep}{RGB}{44,160,44}
\definecolor{cStoryD}{RGB}{255,127,14}
\definecolor{cIter}{HTML}{4A2545}
\definecolor{cDream}{HTML}{7B4FA8}
\pgfplotsset{
  cElect/.style ={color=cElect},
  cConsi/.style ={color=cConsi},
  cOnep/.style  ={color=cOnep},
  cStoryD/.style={color=cStoryD},
  cIter/.style  ={color=cIter},
}

\usepackage{listings}
\usepackage{xcolor}
\lstdefinestyle{promptstyle}{
  basicstyle=\ttfamily\scriptsize,
  breaklines=true,
  breakindent=0pt,
  columns=fullflexible,
  keepspaces=true,
  frame=single,
  framesep=5pt,
  rulecolor=\color{gray!50},
  backgroundcolor=\color{gray!5},
  xleftmargin=3pt,
  xrightmargin=3pt,
  aboveskip=6pt,
  belowskip=6pt,
}

\usepackage[pagebackref,breaklinks,colorlinks]{hyperref}

\usepackage[capitalize]{cleveref}
\crefname{section}{Sec.}{Secs.}
\Crefname{section}{Section}{Sections}
\Crefname{table}{Table}{Tables}
\crefname{table}{Tab.}{Tabs.}

\def\wacvPaperID{3613}
\def\confName{WACV}
\def\confYear{2027}

\begin{document}

\title{ELECT: Settling Identity Before Rendering for Scalable Story Visualization}

\author{%
\begin{tabular}{@{}*{3}{>{\centering\arraybackslash}p{0.31\textwidth}}@{}}
Mehrajul Abadin Miraj$^{*}$ & Md. Mahmudul Hasan       & Meherun Farzana \\[3pt]
Aniket Joarder              & Mahmudul Hasan           & Abir Chakraborty Partha \\[3pt]
Md. Ahasanul Alam           & Redwan Ahmed Rizvee      & Md. Fahim Arefin \\[3pt]
Md Mahmudur Rahman          & Md. Tanvir Alam          & Md Mosaddek Khan
\end{tabular}\\[4pt]
{\normalsize University of Dhaka}\\[2pt]
{\small $^{*}$Corresponding author: \texttt{mehrajulabadin-2020115630@cs.du.ac.bd}}%
}
\maketitle


\begin{abstract}
Training-free story visualization without reference images must preserve recurring entities across frames while scaling to long narratives. Existing methods
typically maintain consistency by conditioning each frame on other frames of
the same story, coupling generation time, VRAM, or both to story length. We
introduce ELECT\footnote{Code and the long-story dataset are provided in the supplementary
material and will be made publicly available upon acceptance.}, which instead settles cross-frame identity before final
rendering through a language-guided scene--entity mapping and stationary
anchors elected from the model's own drafts. This makes the per-frame cost of
the final rendering pass independent of story length. Joint attention exceeds
96 GB of VRAM at 32 scenes, whereas ELECT remains at 8.3\,s per frame and
under 20\,GB VRAM from 10 to 1000 scenes. On the Story50 long-story
benchmark, ELECT reaches 0.833 average Character Consistency Similarity (aCCS),
compared with 0.826 for ten-round refinement, at one seventh of its generation
cost, with lower text alignment (0.327 vs.\ 0.340). At 1000 scenes, to our
knowledge, the first evaluation of training-free story visualization at this
length, ELECT exceeds ten-round refinement by 0.017 aCCS and the nearest
flat-cost baseline by 0.091, while also surpassing refinement in text
alignment at one twenty-fourth of its generation time. These results show that
long-horizon consistency need not require increasingly expensive inter-frame
conditioning as stories grow.
\end{abstract}

\section{Introduction}
\label{sec:intro}
A coherent visual story requires recurring entities to remain consistent
across scenes; otherwise, identity drift breaks narrative continuity even
when individual frames are visually plausible. Story visualization addresses
this challenge by rendering narrative text as a sequence of images while
preserving entity identity across frames. The task was introduced in a
closed-world form by StoryGAN~\citep{storygan}, where the cast is fixed and
seen during training. We study the open-ended form: the cast is unknown until
the text is read, no reference images are provided, and recurring entities may
disappear for long stretches before reappearing later in the story. This
setting better reflects practical story-generation scenarios, where the cast
emerges from the narrative and character-specific visual references are not
assumed to be available. Given only raw prose, a system must determine which
entities recur, establish their visual identities, and preserve those
identities throughout the narrative.

Training-based approaches~\citep{dreambooth,talecrafter,storyimager} bind
identity into model parameters, but admitting a new character then requires
additional optimization, which is poorly suited to open-ended narratives in
which new characters may be encountered throughout the story. 
Moreover,  requiring a photograph or appearance sheet
for each character as a reference would instead shift identity specification to the user, becoming increasingly impractical as the cast grows. We therefore work in the
training-free, reference-free setting, with the prose as the only input.

In this setting, the central technical difficulty is that independently
generated diffusion frames have no persistent channel through which entity
identity can be carried across scenes. A character described in scene $i$ may
be absent for many scenes and return in scene $j$, with $j \gg i$; each frame
follows a separate denoising trajectory from independent noise, and nothing in
the generation of scene $j$ knows what was drawn in scene $i$. Training-free
methods therefore differ mainly in how they construct such a channel.
Autoregressive conditioning~\citep{pan2024arldm,liu2024storygen} passes each
frame to its successor, propagating appearance and errors alike. Attention
sharing~\citep{tewel2024consistory,zhou2024storydiffusion} jointly generates a
batch and extends self-attention across frames, reducing compounding drift at
the cost of keeping the batch resident in VRAM. Story-Iter~\citep{storyiter}
instead regenerates the story over multiple rounds, conditioning each frame on
all frames from the previous round. The mechanisms differ, but the underlying
premise is shared: maintaining consistency requires visual information from
other frames of the same story to remain in the conditioning path during
generation.

This shared premise makes long-horizon consistency increasingly costly in
generation time, VRAM, or both. Under all-pairs conditioning, the $N$
reference image embeddings from previous round form a sequence attended at every denoising step, so one
refinement round incurs $O(N^2)$ attention work and a key/value cache that
grows linearly with $N$; repeated rounds compound these costs. The effect
becomes pronounced as stories grow: Story-Iter's per-frame time increases by a
factor of 3.9 between 10 and 1000 scenes, while ConsiStory's joint attention
exceeds 96 GB of VRAM at 32 scenes (\cref{sec:scaling}). These bottlenecks
arise from the mechanism used to maintain cross-frame consistency, rather than
from any intrinsic requirement of the narrative itself.

To break this coupling, we propose ELECT, a training-free, reference-free
approach that settles cross-frame identity before final rendering rather than
maintaining it through inter-frame visual conditioning. This allows each frame
to be rendered independently, decoupling per-frame
generation cost from story length. ELECT remains competitive with iterative
refinement at benchmark length, while at 1000 scenes it surpasses ten-round
refinement in both identity consistency and text alignment using one
twenty-fourth of the generation time and one third of the VRAM. The relative
advantage therefore grows substantially as the story horizon increases. Our contributions are as follows:
\vspace{-2mm}
\begin{itemize}
    \item ELECT operationalizes pre-render identity settlement through a
    symbolic scene--entity map and a stationary visual anchor for each
    recurring entity, elected from the model's own drafts. This avoids
    both per-character optimization and user-provided visual references
    while enabling independent final-frame generation.
    \item At benchmark length, ELECT reaches 0.833 in average Character
    Consistency Similarity (aCCS), outperforming ten-round refinement by 0.007
    at one seventh of its generation cost, while giving up only 0.013 in
    text alignment. This demonstrates that stationary identity conditioning
    can match or exceed iterative refinement without repeated cross-frame
    generation.
    \item We introduce a long-horizon evaluation suite of 1000-scene stories
    and, to our knowledge, provide the first evaluation of training-free
    story visualization at this scale. ELECT reaches 0.805 aCCS, 0.017
    above ten-round refinement and 0.091 above the nearest flat-cost
    baseline, while maintaining 8.3 s per frame and
19.7 GB VRAM, at \$20 in GPU cost for the suite versus \$487 for
ten-round refinement.
    At this length, refinement takes 55.6 hours, while joint attention
    exhausts VRAM at 32\:scenes.
    \item Through controlled ablations, we identify where ELECT's gains
    arise: anchored conditioning contributes $+0.100$ aCCS at benchmark
    length and $+0.081$ at 1000 scenes; electing the anchor improves aCCS
    by 0.027 over random selection and by 0.100 over the worst candidate,
    while reversing the injection schedule costs 0.048 aCCS.
\end{itemize}

\section{Related Work}
\label{sec:related}

Prior work relevant to ELECT spans open-ended story visualization, methods
for preserving character identity, and training-free mechanisms for
maintaining that identity across frames.

Story visualization began as a closed-world problem. StoryGAN~\citep{storygan}
fixed the cast in advance and showed it during training, and the work that
followed improved the generator or its conditioning without
disturbing that assumption~\citep{maharana,liwordlevel,chen}. StoryGen~\citep{liu2024storygen}
named the open-ended form we study here, in which characters exist only as
descriptions in the prose, and released a large dataset of illustrated
stories spanning hundreds of character categories. Once the cast is
unknown until the text is read, the question becomes where an entity's
appearance is stored and how it reaches a frame that needs it.

One answer stores it in weights. Subject-driven
personalization~\citep{dreambooth,textualinversion,elite,photomaker} learns
a token or a low-rank update per character, and story systems built on it
compose the learned characters into
scenes~\citep{talecrafter,storyimager,diffsensei}. Identity fidelity can be
strong, but admitting a new character costs an optimization run, and
continual customization~\citep{cidm,storyweaver} reports identity degrading
as learned characters accumulate, which is the wrong direction for a
narrative whose cast keeps growing.

The training-free alternative stores appearance in the frames themselves
and manufactures a channel between them. Autoregressive
conditioning~\citep{pan2024arldm,liu2024storygen} hands each frame to its
successor, keeping neighbours consistent while inheriting every error it is
given. Attention sharing~\citep{tewel2024consistory,zhou2024storydiffusion}
generates a batch jointly and keeps it resident in VRAM; StoryDiffusion
additionally bootstraps long stories from its first generated frames, tying
everything later to those initial outputs.
Story-Iter~\citep{storyiter} instead regenerates the story over several
rounds, conditioning each frame on all frames of the previous round
compressed to a fixed number of global tokens, which keeps per-frame
conditioning cheap at the price of repeated full passes over the story
(\cref{sec:scaling}) which is still quadratic. The injection mechanism is largely shared,
IP-Adapter's decoupled cross-attention~\citep{ye2023ipadapter}; what
separates these methods is which visual references reach each frame, and
that choice is what couples their cost to story length.

A second family breaks that coupling by conditioning on the character
rather than the frame, fixing the conditioning before rendering: clustering
samples into an identity, minting a portrait from an LLM description,
sharing one reference panel, or grounding mentions to entity
features~\citep{chosenone,dreamstory,story2board,freestory}; concurrent
multi-shot video work accumulates per-entity banks across shots. ELECT
belongs here, but where these approaches accept the reference obtained at
construction time, we elect it from evidence distributed across the
complete story. A parallel line delegates narrative structure to a language
model~\citep{cheng2024theatergen,autostudio,seedstory,moviedreamer}, and
existing benchmarks~\citep{storybench,freestory,vistorybench} run to tens of
frames, so neither the mechanism nor the evaluation has yet been pushed to
the horizon we target.


Taken together, prior work addresses long horizon
identity consistency through either
per-character optimization or story-length-dependent
inter-frame visual conditioning, which becomes impractical
in generation time and VRAM as stories grow.

\section{The ELECT Framework}
\label{sec:method}
\begin{figure*}[t]
    \centering
    \includegraphics[width=0.92\textwidth]{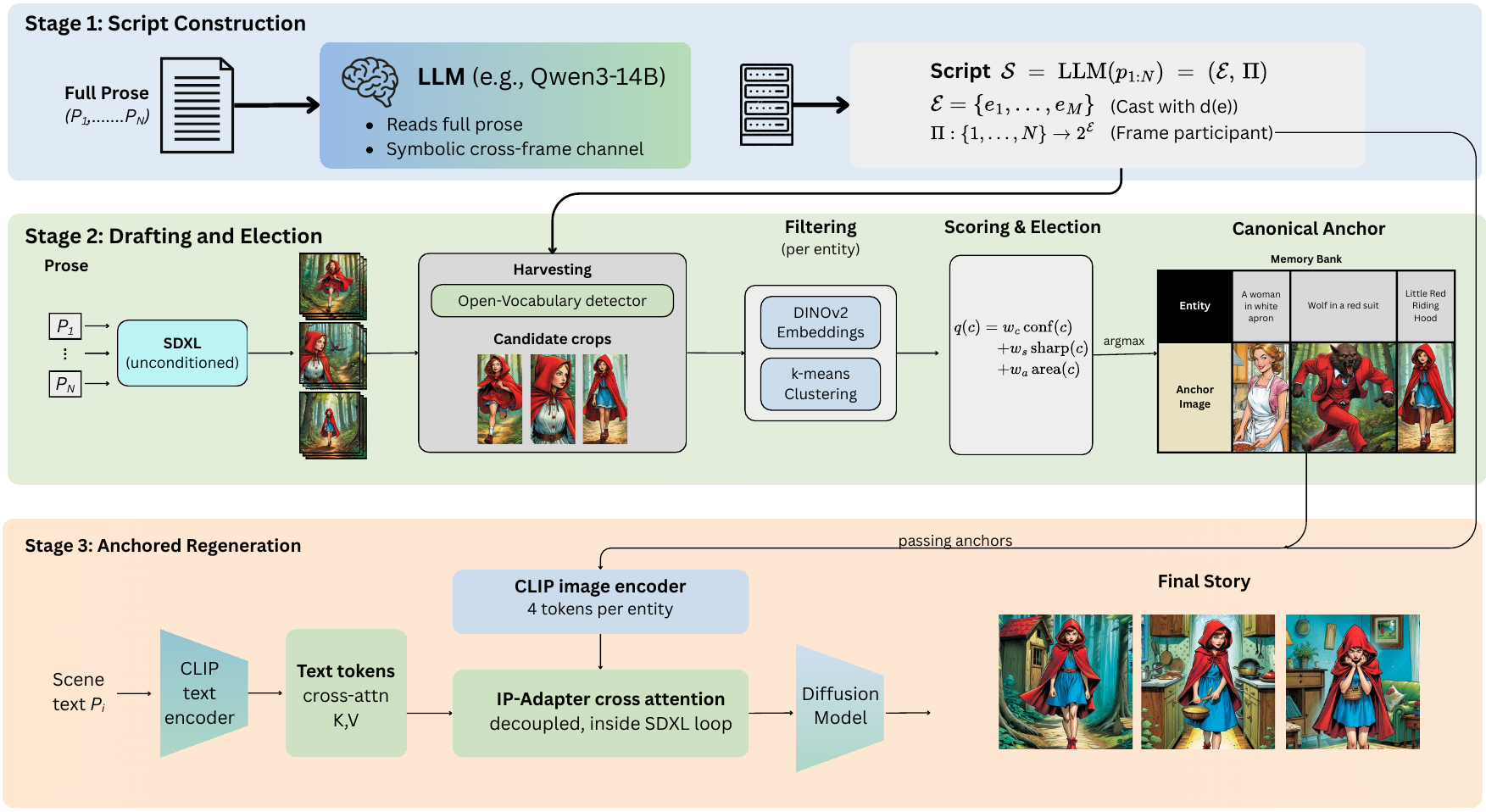}
    \caption{\textbf{Overview of ELECT.} \emph{Stage 1, script construction}: an LLM reads the full story and writes a Script, a recurring cast with short descriptions and a per-frame participant map, establishing cross-frame consistency symbolically. \emph{Stage 2, drafting and election}: SDXL drafts each frame from text alone, a detector proposes crops for that frame's cast, DINOv2 filtering removes misfires, and the best crop per entity is elected as its canonical anchor. \emph{Stage 3, anchored regeneration}: each frame is regenerated from scratch using its own text and its participants' anchors via IP-Adapter, independently of other frames, so the pass stays parallel and flat in $N$.}
    \label{fig:elect}
\end{figure*}

We now describe ELECT, which transforms a sequence of story scenes into a
visually consistent image sequence through three stages: script construction,
drafting and anchor election, and anchored regeneration. Figure~\ref{fig:elect}
provides an overview of the complete pipeline.

\subsection{Problem Setting}
\label{sec:method-setup}
A story is a sequence of $N$ scene descriptions $p_1,\dots,p_N$, one per frame, with no cast list, reference photographs, or guarantee that an entity's first mention provides a usable visual description. We produce frames $x_1,\dots,x_N$ in which recurring entities remain visually consistent, each $x_i$ remains faithful to $p_i$, and the cost of producing $x_i$ depends on neither $i$ nor $N$. Every component is a frozen pretrained model~\citep{sdxl,ye2023ipadapter,dinov2,clip,yoloe} composed at
inference; no weights are updated at any point.

\subsection{Script Construction}
\label{sec:method-script} ELECT begins by constructing a symbolic representation of the story. A local
language model (Qwen3-14B~\cite{yang2025qwen3technicalreport}) reads the full
prompt sequence in one pass and emits a \emph{Script} $\mathcal{S}$.

\begin{equation}
\mathcal{S} \;=\; \mathrm{LLM}(p_{1:N}) \;=\; (\mathcal{E},\,\Pi)
\label{eq:script}
\end{equation}
As defined in \cref{eq:script},
$\mathcal{E}=\{e_1,\dots,e_M\}$ is the cast of recurring entities,
each a character the story returns to, and
$\Pi:\{1,\dots,N\}\rightarrow 2^{\mathcal{E}}$ assigns to each frame the
entities it casts. The model is asked for exactly two things: a cast of
recurring entities, each carrying a one-line description $d(e)$, and, for
every frame, the subset of that cast supported by the frame's own text. The
prompt is reproduced in full in the supplementary material.

Everything required by the later stages is read from this pair. An entity's
description $d(e)$ is used unchanged as the grounding phrase that queries
the detector, while its frame set $F(e)=\{i : e\in\Pi(i)\}$ is obtained by
scanning $\Pi$. Both come directly from $\mathcal{S}$, with no further
judgment asked of the model once it has named the cast and placed its members
scene by scene.

The map $\Pi$ is where the cross-frame channel is established. Other
training-free methods construct it in pixel space, implicitly asserting that
entity $e$ in frames $i$ and $j$ is the same by making their latents attend
to one another; \cref{eq:script} asserts it symbolically. Deciding that
``the knight,'' ``he'', and ``Sir Aldric'' denote one entity in scene $j$
as in scene $i$, with $j \gg i$, is a coreference-resolution problem, handled notably
more reliably in text than in latent space.


\subsection{Drafting and Election}
\label{sec:method-draft}
\label{sec:method-elect}
Each frame is first generated once from its own prompt with a pretrained text-to-image diffusion model~\cite{sdxl} under a fixed style template, with no conditioning on any other frame. These drafts are not shown; instead, they provide the candidate pool from which entity anchors are selected. Anchor construction proceeds in two steps.

\paragraph{Harvesting and filtering.}
An open-vocabulary detector~\citep{yoloe} runs on every draft, queried only with that frame's Script participants, yielding a pool of candidate crops for each entity across the story. Each pool is embedded with DINOv2~\citep{dinov2} and clustered with $k$-means. Crops outside the dominant cluster are discarded, leaving a candidate pool $\mathcal{C}(e)$ for each entity. This filtering also removes two common errors: detections of the wrong object and lookalike characters that match the same grounding phrase. Such incorrect crops appear as outliers in the embedding space.

\paragraph{Election.}
Given the filtered candidate pool $\mathcal{C}(e)$, the anchor for entity
$e$ is selected as the highest-scoring surviving crop,
$a(e)=\arg\max_{c\in\mathcal{C}(e)} q(c)$, where the quality score combines
three per-crop signals,
\begin{equation}
q(c) \;=\; w_c\,\mathrm{conf}(c) \;+\; w_s\,\mathrm{sharp}(c)
           \;+\; w_a\,\mathrm{area}(c),
\label{eq:election}
\end{equation}
with the balanced default $(w_c,w_s,w_a)=(0.4,0.3,0.3)$. The three terms
capture complementary aspects of reference quality.
$\mathrm{conf}(c)$ is the detector's confidence, which is low for partial,
occluded, and weakly grounded instances; $\mathrm{sharp}(c)$ is a
normalised Laplacian response, which is low for blurred and out-of-focus
crops; and $\mathrm{area}(c)$ is the crop's area as a fraction of its
frame, which is low for distant instances that contain too few pixels to
encode appearance reliably. These criteria matter because the elected
anchor is re-encoded and injected into every frame in which its entity
appears, so defects in the selected crop can propagate throughout the story.
\Cref{sec:ablation} varies $(w_c,w_s,w_a)$ and shows that no single signal suffices.

By the time \cref{eq:election} is applied, two earlier decisions have already
been made. Harvesting determines whether a crop corresponds to the target
entity, since a candidate exists only because the detector matched it to the
entity's grounding phrase. Filtering then determines whether that crop is
representative of the entity: restricting $\mathcal{C}(e)$ to the dominant
DINOv2 cluster removes unrepresentative crops as a hard constraint rather
than treating representativeness as a soft penalty traded against image
quality. Election therefore addresses the remaining question: which of the
surviving, matched, and representative crops provides the cleanest instance
to re-encode.

\subsection{Anchored Regeneration}
\label{sec:method-regen} 
With the anchor bank fixed, every frame is regenerated independently from fresh Gaussian noise and its own scene text. Each frame is conditioned through IP-Adapter's decoupled cross attention~\citep{ye2023ipadapter} only
on the bank entries of the entities participating in that scene, with one token block per active entity built from the CLIP embedding of its anchor.
Because the bank entries are entity crops rather than whole frames,
composition and background are not transferred along with identity.
Moreover, because each frame is conditioned only on the entities in its own
scene, the conditioning set does not grow with story length. Because a scene casts a bounded number of entities, what a frame is conditioned on does not grow with the number of frames preceding it, and a 50-scene and a 1000-scene story pay the same per-frame cost.

\paragraph{Injection schedule.}
The remaining design choice is how strongly, and when, the anchor should
influence generation. We scale the anchor token block by a factor
$\lambda_{\mathrm{ent}}(t)$ inside the decoupled cross-attention and vary it
linearly over the sampling trajectory,
\begin{equation}
\lambda_{\mathrm{ent}}(t) = \lambda_{\mathrm{hi}}
  + \big(\lambda_{\mathrm{lo}}-\lambda_{\mathrm{hi}}\big)\,
    \frac{T-t}{T},
\label{eq:schedule}
\end{equation}
with $t=T,\dots,0$ over the $T$ sampling steps, from $\lambda_{\mathrm{hi}}=0.85$ at the first, highest-noise step to
$\lambda_{\mathrm{lo}}=0.35$ at the last. Early steps determine coarse
structure, where silhouette, build, and palette are established and weak
anchor guidance can lose identity; later steps determine texture and local
detail, where reducing anchor strength leaves more capacity for the scene
text. \cref{tab:strength} compares this decay with constant scales and the reverse sweep.


\section{Experiments}
\label{sec:experiments}
We evaluate ELECT along the dimensions that motivate its design: identity
consistency and text alignment, scalability in generation time and VRAM,
and the contribution of its individual components. We first describe the
evaluation setting and implementation.
\subsection{Setup}
\label{sec:setup}
We evaluate on the Story50 long-story benchmark (20 stories of 50 frames
each)~\cite{storyiter} and on five 1000-scene stories adapted from
public-domain books: \emph{The Mysterious Island}, \emph{Taras Bulba}, the
\emph{Shahnameh}, a Russian fairy tale, and the \emph{V\"olsunga} saga.
Their construction is described in the supplementary material. We report
CLIP-T for text--image alignment and aCCS for cross-frame identity
consistency, following~\cite{cheng2024theatergen}.


\paragraph{Implementation details.}
\label{sec:impl}
All runs use SDXL~\citep{sdxl} at $1024\times1024$ with 50 sampling steps
on a single NVIDIA RTX 6000 PRO Blackwell (96\,GB). Anchors are injected
through base IP-Adapter with four pooled tokens per entity using the
schedule in \cref{eq:schedule}. Script extraction uses Qwen3-14B.
Harvesting uses YOLOE~\cite{yoloe}, and candidate crops are embedded with
DINOv2~\cite{dinov2} for the cluster filter in \cref{sec:method-elect}.
In preliminary experiments, YOLOE grounded the Script descriptions more
accurately than the alternatives tested, while DINOv2 separated visually
similar characters more reliably than CLIP

\begin{table}[t]
\centering
\captionsetup{skip=2pt}
\caption{Story50~\cite{storyiter} benchmark comparison.
Rows below the rule are our own reproductions, run on identical stories,
seeds and hardware; rows above are quoted from prior publications.}
\label{tab:benchmarks}
\small
\setlength{\tabcolsep}{4pt}
\renewcommand{\arraystretch}{0.9}
\begin{tabular}{lcc}
\toprule
Method & CLIP-T $\uparrow$ & aCCS $\uparrow$ \\
\midrule
AR-LDM~\citep{pan2024arldm}                     & 0.216 & 0.673 \\
StoryGen~\citep{liu2024storygen}                & 0.223 & 0.740 \\
IP-Adapter~\citep{ye2023ipadapter}              & 0.274 & 0.751 \\
FLUX.1.Kontext~\citep{batifol2025flux}          & 0.290 & 0.783 \\
Nano Banana                                     & 0.310 & 0.791 \\
IP-AdapterXL~\citep{ye2023ipadapter}            & 0.297 & 0.787 \\
DALL$\cdot$E 3~\citep{betker2023dalle3}         & 0.301 & 0.674 \\
SEED-LLaMA~\citep{ge2023seedllama}              & 0.279 & 0.688 \\
ConsiStory~\citep{tewel2024consistory}          & 0.316 & 0.761 \\
\midrule
1Prompt1Story~\citep{liu2025oneprompt}          & 0.340 & 0.7287 \\
StoryDiffusion~\citep{zhou2024storydiffusion}   & \textbf{0.351} & 0.8103 \\
DreamStory~\cite{dreamstory}  & 0.3193 & 0.7627 \\
Story-Iter~\cite{storyiter}                         & 0.340 & 0.8264 \\
\textbf{ELECT (ours)}                           & 0.3273 & \textbf{0.8334} \\
\bottomrule
\end{tabular}
\end{table}

\subsection{What does anchoring buy?}
\label{sec:quality}

\paragraph{Anchoring drives consistency.}
Anchoring replaces per-frame identity guesswork with a single settled
reference: Pass~0 drafts every scene independently from text, one crop per entity is
elected as the anchor, and the second pass conditions all remaining scenes on
it according to the mapping. On the standard story50 benchmark (\cref{tab:benchmarks}), this puts
ELECT at $0.8334$ aCCS, the highest of any method we compare against, ahead of
Story-Iter ($0.8264$) and StoryDiffusion ($0.8103$), while CLIP-T sits at
$0.3273$, within $0.024$ of the best value and above most
baselines. The benchmark therefore places us where anchoring should: identity
is held tighter than by iterative refinement or reference-injection methods, at
a small and expected cost in per-frame prompt adherence, since conditioning on
a fixed appearance necessarily pulls each frame slightly away from its own
text.

Pass 0 isolates the anchor's contribution: it shares prompts, backbone, style template and seeds with the final pass and differs only in that no anchor is injected (\cref{tab:pass0}). Anchoring raises aCCS by $0.100$ at 50 scenes and by $0.081$ at 1000, while CLIP-T falls by only $0.025$ and $0.012$. The trade is strongly asymmetric, and it does not degrade with story length.

\begin{table}[t]
\centering
\caption{Effect of anchored conditioning. Pass~0 is ELECT's own draft:
identical prompts, backbone and seeds, generated independently from text
with no cross-frame conditioning. The 50-scene row is the mean over 20
benchmark stories.}
\label{tab:pass0}
\small
\setlength{\tabcolsep}{5pt}
\begin{tabular}{l cc cc}
\toprule
& \multicolumn{2}{c}{aCCS $\uparrow$} & \multicolumn{2}{c}{CLIP-T $\uparrow$} \\
\cmidrule(lr){2-3}\cmidrule(lr){4-5}
& Pass 0 & ELECT & Pass 0 & ELECT \\
\midrule
\multicolumn{5}{l}{\emph{50-scene benchmark}} \\
Mean over 20 stories & 0.7334 & \textbf{0.8334} & \textbf{0.352} & 0.3273 \\
$\Delta$ & \multicolumn{2}{c}{$+0.100$} & \multicolumn{2}{c}{$-0.025$} \\
\midrule
\multicolumn{5}{l}{\emph{1000 scenes}} \\
Mean & 0.7239 & \textbf{0.8050} & \textbf{0.318} & 0.306 \\
$\Delta$ & \multicolumn{2}{c}{$+0.081$} & \multicolumn{2}{c}{$-0.012$} \\
\bottomrule
\end{tabular}
\end{table}

\subsection{What does story length cost?}
\label{sec:scaling}
\label{sec:long}

\begin{table}[t]
\centering
\caption{Quality and cost at $N{=}1000$ scenes ($1024^2$, RTX 6000 PRO Blackwell, 96\,GB). ConsiStory exceeds VRAM at 32 scenes; its peak is
reported at $N{=}16$. \textbf{s/frame is end-to-end wall clock divided by
$N$}, covering every stage each method runs: for ELECT, Script extraction,
the Pass~0 drafts, detection, embedding and clustering, and the anchored
regeneration pass; for Story-Iter, all ten refinement rounds. Total\,(h) is
this average per story generation time CLIP-T and aCCS are mean over 5 stories.}
\label{tab:n1000}
\setlength{\tabcolsep}{3pt}
\renewcommand{\arraystretch}{0.9}
\footnotesize
\begin{tabular}{@{}lccccc@{}}
\toprule
Method & CLIP-T $\uparrow$ & aCCS $\uparrow$ & s/frame & Total (h) & VRAM (GB) \\
\midrule
\multicolumn{6}{@{}l}{\emph{Cost grows with $N$}} \\
ConsiStory~\citep{tewel2024consistory}        & OOM & OOM & -- & -- & 92.7 ($N{=}16$) \\
Story-Iter~\citep{storyiter}                  & 0.3044 & 0.7885 & 200.1 & 55.6 & 59.7 \\
\midrule
\multicolumn{6}{@{}l}{\emph{Cost flat in $N$}} \\
1Prompt1Story~\citep{liu2025oneprompt}        & 0.309 & 0.714 & 7.37 & 2.05 & \textbf{16.7} \\
StoryDiffusion~\citep{zhou2024storydiffusion} & \textbf{0.315} & 0.639 & \textbf{4.40} & \textbf{1.22} & 66.4 \\
DreamStory~\cite{dreamstory} & 0.3089 & 0.6976 & 38.42 & 10.7 & 32.9 \\
\textbf{ELECT (ours)}                          & 0.306 & \textbf{0.8050} & 8.30 & 2.30 & 19.7 \\
\bottomrule
\end{tabular}
\end{table}

\begin{figure}[h]
\centering
\resizebox{\columnwidth}{!}{%
\begin{tikzpicture}
\begin{groupplot}[
group style={
group size=1 by 2,
vertical sep=9pt,
xticklabels at=edge bottom,
xlabels at=edge bottom,
},
width=7.9cm,
xmin=-25, xmax=1055,
xtick={0,100,200,300,400,500,600,700,800,900,1000},
xticklabels={0,100,200,300,400,500,600,700,800,900,1000},
minor xtick={50,150,250,350,450,550,650,750,850,950},
xlabel={Story length $N$ (frames)},
grid=both, grid style={gray!16}, minor grid style={gray!8},
tick label style={font=\scriptsize},
label style={font=\scriptsize},
every axis plot/.append style={line width=0.9pt, mark size=1.5pt},
]

\nextgroupplot[
height=4.1cm,
ymin=0, ymax=215,
ytick={0,50,100,150,200},
minor ytick={25,75,125,175},
ylabel={Time / frame (s)},
legend style={at={(0.5,1.03)}, anchor=south, legend columns=3,
draw=gray!45, fill=white, font=\scriptsize,
inner sep=2pt, column sep=5pt},
legend cell align=left,
]
\addplot[name path=EL, cElect, mark=*] coordinates {
(10,8.30) (50,8.30) (100,8.30) (250,8.30) (500,8.30) (1000,8.30)};
\addlegendentry{ELECT (ours)}
\addplot[name path=SI, cIter, mark=o, dashed] coordinates {
(10,51.79) (50,56.68) (100,62.61) (150,68.65) (200,75.70)
(250,82.07) (450,111.43) (550,130.78) (1000,200.15)};
\addlegendentry{Story-Iter}
\addplot[cIter!16, opacity=0.55, forget plot] fill between[of=SI and EL];
\addplot[cConsi, mark=triangle*] coordinates {(4,11.22) (8,11.02) (16,11.21)};
\addlegendentry{ConsiStory}
\addplot[cOnep, mark=square*] coordinates {
(4,11.11) (8,9.20) (16,8.38) (32,7.87) (250,7.37) (1000,7.37)};
\addlegendentry{1Prompt1Story}
\addplot[cStoryD, mark=diamond*] coordinates {
(4,4.14) (8,4.13) (16,4.48) (32,4.80) (64,5.13) (256,3.54) (1000,4.40)};
\addlegendentry{StoryDiffusion}
\addplot[cDream, mark=pentagon*] coordinates {
(10,38.42) (50,38.42) (100,38.42) (250,38.42) (500,38.42) (1000,38.42)};
\addlegendentry{DreamStory}
\addplot[cConsi, only marks, mark=x, mark size=3pt, line width=1.1pt, forget plot]
coordinates {(32,11.21)};
\node[font=\tiny, cIter, anchor=center, inner sep=1pt] at (axis cs:600,55) {$24\times$ at $N{=}1000$};
\node[font=\tiny, cElect, anchor=west, inner sep=1pt] at (axis cs:120,26) {ELECT: constant $8.3$\,s/frame};
\draw[->, cElect, line width=0.6pt] (axis cs:180,22) -- (axis cs:210,11.5);

\nextgroupplot[
height=3.1cm,
ylabel={Peak VRAM (GB)},
ymin=0, ymax=125,
ytick={0,20,40,60,80,100},
minor ytick={10,30,50,70,90},
]
\fill[red!7] (axis cs:-25,95.6) rectangle (axis cs:1055,125);
\addplot[black, dotted, line width=0.9pt, forget plot]
coordinates {(-25,95.6) (1055,95.6)};
\node[font=\tiny, anchor=north east, inner sep=1.5pt] at (axis cs:1040,93) {GPU limit $96$\,GB};
\addplot[cElect, mark=*] coordinates {
(10,19.65) (50,19.65) (100,19.65) (250,19.65) (500,19.65) (1000,19.65)};
\addplot[cIter, mark=o, dashed] coordinates {
(10,20.29) (50,20.29) (100,20.29) (150,22.29) (200,23.74)
(250,26.54) (450,35.52) (550,42.33) (1000,59.67)};
\addplot[cConsi, mark=triangle*] coordinates {(4,38.64) (8,79.76) (16,92.71)};
\addplot[cOnep, mark=square*] coordinates {
(4,16.70) (8,16.66) (16,16.70) (32,16.64) (250,16.66) (1000,16.66)};
\addplot[cStoryD, mark=diamond*] coordinates {
(4,66.43) (8,66.43) (16,66.43) (32,66.43) (64,66.43) (256,66.43) (1000,66.43)};
\addplot[cDream, mark=pentagon*] coordinates {
(10,32.90) (50,32.90) (100,32.90) (250,32.90) (500,32.90) (1000,32.90)};
\addplot[cConsi, only marks, mark=x, mark size=3pt, line width=1.1pt, forget plot]
coordinates {(32,92.71)};
\node[font=\tiny, cConsi, anchor=west, inner sep=1.5pt] at (axis cs:95,107) {ConsiStory OOM ($N{\ge}32$)};
\draw[cConsi, line width=0.6pt] (axis cs:90,105) -- (axis cs:38,96);
\end{groupplot}
\end{tikzpicture}%
}
\caption{\textbf{Scaling from 4 to 1000 scenes.} \emph{Top:} amortized cost per
frame. ELECT is constant at $8.3$\,s/frame while Story-Iter degrades from
$51.8$ to $200.1$\,s/frame, a $24\times$ gap at $N{=}1000$ ($8.3$\,s vs.
$200$\,s end-to-end); the shaded band is the resulting speedup.
\emph{Bottom:} peak VRAM. ELECT is constant at $19.7$\,GB, whereas ConsiStory
exceeds the $96$\,GB device budget at $N{=}32$ (cross), while StoryDiffusion and
DreamStory hold fixed budgets of $66.4$ and $32.9$\,GB. Baselines limited to
short horizons are plotted only over the range they complete.}
\label{fig:scaling}
\end{figure}
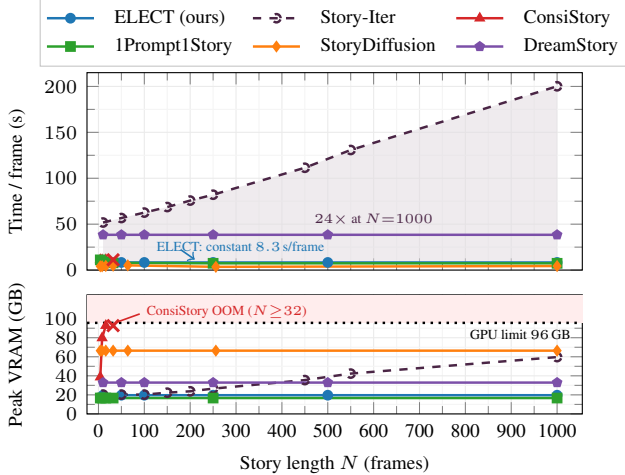

Existing benchmarks typically cover only tens of frames, leaving the
long-story regime largely unexplored. We therefore evaluate stories up to
1000 scenes to examine how quality and computational cost scale with
narrative length. \Cref{fig:scaling} reports scaling from 4 to 1000 scenes,
while \cref{tab:n1000} summarizes quality and cost at 1000 scenes.

\paragraph{ELECT keeps its cost fixed as stories grow.}
The difference is clearest in per-frame cost. ConsiStory cannot reach 32
scenes because its shared attention state exceeds 96 GB of VRAM
(\cref{fig:scaling}). Story-Iter reaches 1000 scenes, but its per-frame cost
grows from $51.8$\,s at 10 scenes to $200.1$\,s at 1000, requiring 55.6
hours for one story. ELECT instead remains at $8.3$\,s per frame and
$19.7$\,GB VRAM from 10 to 1000 scenes, completing the same story in
2.30 hours.

This behavior follows from how ELECT carries cross-frame information. The
Script and frozen anchor bank fix the relevant entity information before
rendering, so each scene is conditioned only on the entities it contains and
the conditioning size does not grow with story length. 1Prompt1Story and
StoryDiffusion also maintain nearly constant per-frame cost and are faster
than ELECT at 1000 scenes, at $7.37$\,s and $4.40$\,s per frame,
respectively. However, this speed advantage comes with substantially lower
identity consistency: ELECT reaches $0.8050$ aCCS, compared with $0.714$ for
1Prompt1Story and $0.639$ for StoryDiffusion. Thus, ELECT retains markedly
stronger identity consistency while keeping its per-frame cost nearly
constant as stories grow.

\paragraph{The efficiency does not come at the cost of identity.}
At 1000 scenes, ELECT reaches $0.8050$ aCCS, which is $0.091$ higher than
1Prompt1Story, $0.166$ higher than StoryDiffusion, and $0.017$ higher than
Story-Iter (\cref{tab:n1000}). CLIP-T is similar across the three
flat-cost methods, with ELECT only $0.009$ below StoryDiffusion. Thus,
the methods that keep their cost fixed do not all preserve identity equally
well. StoryDiffusion, for example, is only $0.023$ behind ELECT at the
benchmark length but falls $0.166$ behind at 1000 scenes. Against
Story-Iter, ELECT is better on both quality measures at 1000 scenes while
using about $24\times$ less generation time and one-third the peak VRAM.


\subsection{How hard should the anchor push, and which crop should it push with?}
\label{sec:ablation}

Results in \cref{tab:pass0} already show that anchoring improves identity
consistency. We next examine two choices that determine how much of this gain
is realised: how the anchor is injected during sampling
(\cref{tab:strength}) and which crop is elected as the anchor
(\cref{tab:election}).

\paragraph{When the anchor pushes matters more than how hard.}
At a constant scale, increasing $\lambda_{\mathrm{ent}}$ from $0.35$ to
$0.85$ raises identity monotonically, from 0.787 to 0.829 aCCS, while
reducing alignment from 0.343 to 0.327 CLIP-T. A fixed scale therefore acts
largely as a dial between the two objectives.

The scheduled variants reveal
that timing also matters. Both use the same endpoints and differ only in
direction, yet they finish $0.048$ aCCS apart: decaying
$0.85\!\to\!0.35$ is the best arm in the table, while rising
$0.35\!\to\!0.85$ is the worst. Because the two schedules have the same
average anchor strength, the difference comes from when that strength is
applied. The rising schedule performs no better on either metric than
holding the weakest scale throughout, consistent with identity-relevant
structure being established early in denoising. Conversely, decay
outperforms a constant $0.85$ on both metrics despite using lower strength
for most of the trajectory, suggesting that strong early guidance and weaker
late guidance provide a better balance. The flat schedules remain useful for
other operating points: a constant $0.50$ reaches $0.808$ aCCS and $0.338$
CLIP-T, matching our Story-Iter reproduction's alignment to within $0.002$
at $0.018$ lower consistency.



\paragraph{Anchor election is necessary, not merely anchoring.}
To separate having an anchor from having a good one, we cross the election
rule with the pool it draws from: argmax of \cref{eq:election}, a random
survivor, or argmin, over either the cluster-filtered pool or the raw one
(\cref{tab:election}). aCCS is monotonic in the rule on both pools, $0.8334
/ 0.8065 / 0.7911$ filtered and $0.8244 / 0.7951 / 0.7335$ raw, and the best
cell requires both stages: filtering without a careful election, or election
over an unfiltered pool, each falls short of the two together. A random
anchor already recovers much of the benefit of anchoring, but a poor one
gives most of it back, especially in the raw pool, where argmin falls
$0.091$ below argmax.

The two stages guard against different failures. Filtering helps every rule,
but by $0.009$ aCCS for argmax and $0.011$ for random against $0.058$ for
argmin, so its work is almost entirely the removal of unusable candidates,
while the election decides which of the survivors is actually re-encoded.
Alignment is essentially unaffected by the choice: argmax and random read
$0.3273$ and $0.3280$ CLIP-T on the filtered pool. The identity gain
therefore comes from choosing the anchor well, at no meaningful cost in
alignment.
\paragraph{Combining the three signals is better than using any one alone.}
We ablate the three terms in the election score by comparing the balanced
configuration with each single-signal variant, keeping all other settings
fixed (\cref{tab:arms}). The balanced weights achieve the highest aCCS
(0.8334), outperforming area only (0.8291), confidence only (0.8118), and
sharpness only (0.7980). This shows that no single signal is sufficient to
select the best anchor, while combining confidence, sharpness, and area
gives the strongest overall selection. CLIP-T follows the same trend
except for area only, which is marginally higher than the balanced setting
(0.3279 vs.\ 0.3273). We therefore use the balanced weights
$(0.4,0.3,0.3)$ throughout.

\begin{table}[t]
\centering
\caption{Anchor strength during regeneration.
$\lambda_{\mathrm{ent}}$ scales the anchor token block inside IP-Adapter's
decoupled cross-attention, either held flat across the 50 sampling steps
or swept linearly by \cref{eq:schedule}. The two scheduled arms share
their endpoints and differ only in direction. Mean over the story50
stories; everything else fixed, including Script, drafts, harvest,
election and seed. The last row is the configuration reported throughout.}
\label{tab:strength}
\small
\setlength{\tabcolsep}{5pt}
\begin{tabular}{lcc}
\toprule
$\lambda_{\mathrm{ent}}$ & aCCS $\uparrow$ & CLIP-T $\uparrow$ \\
\midrule
\multicolumn{3}{l}{\emph{Flat}} \\
\quad $0.35$                              & 0.7872 & \textbf{0.3427} \\
\quad $0.50$                              & 0.8083 & 0.3382 \\
\quad $0.75$                              & 0.8107 & 0.3300 \\
\quad $0.85$                              & 0.8293 & 0.3268 \\
\midrule
\multicolumn{3}{l}{\emph{Scheduled}} \\
\quad rising $0.35\!\rightarrow\!0.85$    & 0.7850 & 0.3426 \\
\quad decaying $0.85\!\rightarrow\!0.35$  & \textbf{0.8334} & 0.3273 \\
\bottomrule
\end{tabular}
\end{table}

\begin{table}[t]
\centering
\caption{Where the anchor's quality comes from: the rule that picks one
crop per entity, crossed with whether the pool it picks from has been
cluster-filtered. All cells share the pipeline, prompts, backbone and
seeds, and differ only in these two factors. \emph{Random} is the mean of
three seeds, whose spread is at most 0.009 aCCS. The top-left cell is the
reported configuration of \cref{tab:benchmarks}. Mean over the 20
benchmark stories.}
\label{tab:election}
\small
\setlength{\tabcolsep}{5pt}
\begin{tabular}{lcccc}
\toprule
& \multicolumn{2}{c}{aCCS $\uparrow$} & \multicolumn{2}{c}{CLIP-T $\uparrow$} \\
\cmidrule(lr){2-3}\cmidrule(lr){4-5}
Selection rule & Filtered & Raw pool & Filtered & Raw pool \\
\midrule
argmax of \cref{eq:election} & \textbf{0.8334} & 0.8244 & 0.3273 & 0.3260 \\
Random survivor              & 0.8065 & 0.7951 & \textbf{0.3280} & 0.3226 \\
argmin of \cref{eq:election} & 0.7911 & 0.7335 & 0.3206 & 0.3136 \\
\bottomrule
\end{tabular}
\end{table}

\begin{table}[t]
\centering
\caption{Weights of the election score on the story50 benchmark, driving
$(w_c,w_s,w_a)$ of \cref{eq:election} to each corner of the simplex with
everything else fixed. The first row is the reported configuration.}
\label{tab:arms}
\small
\setlength{\tabcolsep}{5pt}
\begin{tabular}{lcc}
\toprule
$(w_c,w_s,w_a)$ & aCCS $\uparrow$ & CLIP-T $\uparrow$ \\
\midrule
balanced $(0.4,0.3,0.3)$  & \textbf{0.8334} & 0.3273 \\
confidence only $(1,0,0)$ & 0.8118 & 0.3262 \\
sharpness only $(0,1,0)$  & 0.7980 & 0.3243 \\
area only $(0,0,1)$       & 0.8291 & \textbf{0.3279} \\
\bottomrule
\end{tabular}
\end{table}

\subsection{Can Stage 3 reuse the draft's trajectory?}
Pass~0 achieves higher CLIP-T (\cref{tab:pass0}), suggesting that re-entering
its trajectory while supplying identity through the anchor might retain both
alignment and consistency. We test this following SDEdit~\citep{meng2022sdeditguidedimagesynthesis}:
the draft is forward-noised to a start index determined by strength $s$, and
only the remaining steps are run under injection ($s{=}0.75$ skips 13 of 50
steps; our default is $s{=}1$). The resulting trade-off is less favorable
(\cref{tab:noisereuse}). Alignment improves because composition is established
by a pass with no competition from anchor tokens, and the variant preserves
that decision. Identity degrades for the same reason: diffusion commits
low-frequency structure early, precisely where \cref{eq:schedule} assigns
$\lambda_{\mathrm{ent}}$ its highest values. Entering mid-trajectory leaves
the anchor only the low-noise tail, where it can restyle surfaces but not
re-shape a participant already established by the draft. The ordering in
\cref{tab:strength} remains unchanged, showing that the schedule effect is
not an artefact of starting from noise.

\begin{table}[h]
\centering
\captionsetup{skip=2pt}
\caption{Draft-trajectory reuse ($s{=}0.75$) instead of fresh noise;
story50 benchmark. Against the fresh-noise default
(0.8334 aCCS, 0.3273 CLIP-T), reuse costs 0.039 aCCS for 0.017 CLIP-T.}
\label{tab:noisereuse}
\small
\setlength{\tabcolsep}{5pt}
\renewcommand{\arraystretch}{0.9}
\begin{tabular}{lcc}
\toprule
$\lambda_{\mathrm{ent}}$ & aCCS $\uparrow$ & CLIP-T $\uparrow$ \\
\midrule
flat $0.15$                        & 0.7626 & \textbf{0.3512} \\
flat $0.50$                        & 0.7807 & 0.3478 \\
flat $0.70$                        & 0.7867 & 0.3457 \\
decaying $0.85\!\rightarrow\!0.35$ & \textbf{0.7947} & 0.3442 \\
\bottomrule
\end{tabular}
\end{table}

\subsection{Is the planning layer a bottleneck?}
\label{sec:roster}
If the Script has to come from a strong reader, ELECT inherits the cost and deployment constraints of one. We therefore replace Qwen3-14B with Qwen3-4B, a quarter of the parameters, and re-run the pipeline with everything else fixed. The smaller planner costs $0.0118$ aCCS, $0.8216$ against $0.8334$
(\cref{tab:roster}), and moves alignment a negligible $0.0007$ the other
way. Since nothing downstream of the Script changes, the whole difference
lies in what the two models write: which entities they promote to the cast,
which frames they credit them to, and how usable each one-line description
is as a grounding phrase for the detector. A small model recovers about $99\%$ of the reported
score, so the symbolic layer does not depend on frontier-scale reading. It
is not free either: the penalty exceeds the $0.007$ by which ELECT leads
Story-Iter at this length, so the planner is material to that
comparison, though it stays small against the $0.100$ anchoring supplies and
the $0.048$ the schedule's direction is worth (\cref{tab:strength}).

\begin{table}[h]
\centering
\caption{Planner capacity. The Script is re-extracted with a smaller
language model and the pipeline re-run end to end on the 50-scene
benchmark, everything else fixed. The first row is the configuration
reported throughout.}
\label{tab:roster}
\small
\setlength{\tabcolsep}{5pt}
\begin{tabular}{lcc}
\toprule
Script extractor & aCCS $\uparrow$ & CLIP-T $\uparrow$ \\
\midrule
Qwen3-14B & \textbf{0.8334} & 0.3273 \\
Qwen3-4B  & 0.8216 & \textbf{0.3280} \\
\bottomrule
\end{tabular}
\end{table}

\subsection{What does it cost to run?}
\label{sec:cost}

Time and VRAM capture architectural efficiency, but practical deployment
also depends on generation cost. At \$1.74 per GPU-hour (Vast.ai rate for
an NVIDIA RTX 6000 PRO Blackwell), the difference is small at benchmark
scale but compounds sharply with length: on the long-story suite (5 stories
$\times$ 1000 frames), ELECT costs \$20 versus \$487 for ten-round
refinement (\$4.01 vs.\ \$97.44 per story), with the gap widening from
$6.7\times$ to $24\times$ because refinement's per-frame cost grows with
$N$ while ELECT's does not. Scaled to 100 thousand-scene stories, this is
\$401 versus \$9{,}744. ELECT is not the cheapest method (StoryDiffusion:
\$10.63; 1Prompt1Story: \$17.82 for the suite), but it is the cheapest that
preserves identity at this length, with both cheaper alternatives scoring
lower on aCCS (by 0.166 and 0.091, \cref{tab:cost}).

\begin{table}[t]
\centering
\caption{Cost of generating each evaluation set, at \$1.74 per GPU-hour
(NVIDIA RTX 6000 PRO, on-demand(vast.ai), 22 August 2026). Story50 is 20 stories of 50
frames; the long-story suite is five stories of 1000 frames. Totals are
the generation times of \cref{tab:n1000} and \cref{fig:scaling}; hourly
rates and hardware availability vary, so the ratios matter more than the
absolute figures.}
\label{tab:cost}
\small
\setlength{\tabcolsep}{4.5pt}
\begin{tabular}{lcccc}
\toprule
& \multicolumn{2}{c}{Story50} & \multicolumn{2}{c}{Long-story suite} \\
\cmidrule(lr){2-3}\cmidrule(lr){4-5}
Method & Hours & USD & Hours & USD \\
\midrule
Story-Iter~\citep{storyiter}                  & 15.74 & 27.39 & 280.0 & 487.20 \\
1Prompt1Story~\citep{liu2025oneprompt}        & 2.16 & 3.76 & 10.24 & 17.82 \\
DreamStory~\cite{dreamstory} 10.99 & 19.123 & 53.5 & 93.09 \\
StoryDiffusion~\citep{zhou2024storydiffusion} & \textbf{1.22} & \textbf{2.12} & \textbf{6.11} & \textbf{10.63} \\

\textbf{ELECT (ours)}                         & 2.35 & 4.09 & 11.53 & 20.06 \\
\bottomrule
\end{tabular}
\end{table}

\begin{figure}[t]
    \centering
    \includegraphics[width=\columnwidth]{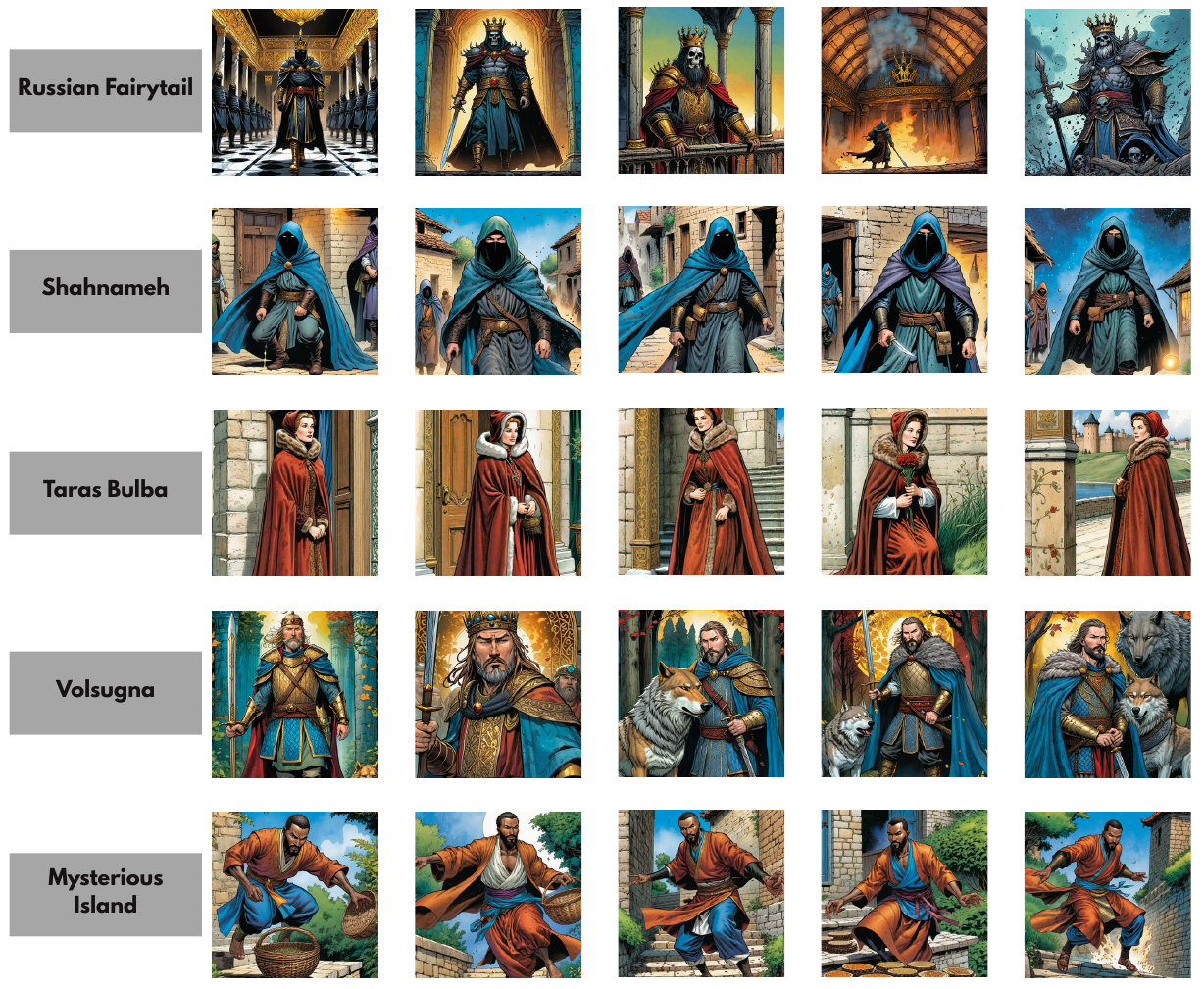}
    \caption{\textbf{One recurring entity per 1000-scene story, shown at five frames
sampled across the full length for the five stories.}
    }
    \label{fig:sample}
\end{figure}

\subsection{What this looks like at 1000 scenes}
\Cref{fig:sample} tracks a single recurring entity in each long-story
suite, at frames drawn from across the whole 1000, rather than from one
neighbourhood. Silhouette, palette, and costume survive the span even when
hundreds of scenes separate two appearances, which is what the flat per-frame
cost protects: the anchor is the same crop at frame $j$ as at frame $i$, with
$j \gg i$, so there is no accumulated drift. It is not exact. Garment detail,
facial features, and accessories vary between appearances, and the $0.8050$
aCCS of \cref{tab:n1000} is the quantitative form of that gap. What the
anchor fixes is the entity's identity at recognition level, not its appearance
pixel for pixel.

\section{Conclusions}
\label{sec:conclusion}
Training-free story visualization has typically maintained consistency
through visual dependencies between frames, making long-horizon generation
increasingly expensive. ELECT instead separates entity correspondence from
appearance: a language model resolves recurring entities from the prose,
while one elected visual anchor per entity fixes appearance before final
rendering. The resulting frames can be generated independently, with
per-frame cost determined by the cast of a scene rather than the length of
the story. At benchmark length, ELECT reaches higher identity consistency
than ten-round refinement at one seventh of its generation cost, while
trading some text alignment. At 1000 scenes, it exceeds refinement by
0.017 aCCS and 0.002 CLIP-T while using one twenty-fourth of its generation
time and one third of its VRAM. Controlled ablations further show that
anchored conditioning provides the dominant consistency gain, while anchor
selection and injection timing materially affect the final result. These
findings suggest that long-horizon consistency need not rely on increasingly
expensive inter-frame conditioning, opening a path toward scalable visual
story\:generation.
{\small
\bibliographystyle{ieee_fullname}
\bibliography{egbib}
}

\end{document}